\documentclass[runningheads]{llncs}
\usepackage{graphicx}
\usepackage{amsmath,amssymb}
\usepackage{url}
\usepackage{hyperref}
\usepackage{textcomp}
\begin{document}
\title{ClimbLab: MATLAB Simulation Platform for Legged Climbing Robotics}
\author{Kentaro Uno\inst{1} \and Warley F. R. Ribeiro\inst{1} \and Yusuke Koizumi\inst{1} \and Keigo Haji\inst{1} \and Koki Kurihara\inst{1} \and William Jones\inst{2} \and Kazuya Yoshida\inst{1}}
\authorrunning{K. Uno et al.}
\institute{Space Robotics Lab (SRL), Department of Aerospace Engineering, Tohoku University, Aoba 6-6-01, Aramaki-Aza, Aoba-ku, Sendai, Miyagi 980-8579, Japan\\
\url{http://www.astro.mech.tohoku.ac.jp/e/}
\and VisionSpace Technologies GmbH, Robert-Bosch Stra\ss e 7, 64293 Darmstadt, Germany\\
\url{https://www.visionspace.com/}}
\maketitle
\begin{center}\small $^*$Corresponding author. $^{**}$Kentaro Uno and Warley F. R. Ribeiro contributed equally to this work.\end{center}
\begin{abstract}
This paper presents an open-sourced MATLAB simulation and analysis platform dedicated to legged climbing robots. This simulator enables the design of any limbed robotic system as an articulated multi-body with a floating base and simulates it walking and climbing in an arbitrary environment. The main variable environmental parameters are inclination, gravity, and ground stiffness, and any point cloud can be installed as the terrain map. Furthermore, the simulator employs a rigid body dynamics engine. This paper first describes the simulator structure, and the computational flow and next presents the representative simulation examples where quadrupedal robots assumed gripping on the wall or climbing on the steep slope.
\keywords{Open-sourced simulator \and Climbing robots \and Legged robots}
\end{abstract}

\section{Introduction}
To date, legged robots have been showing more usefulness in many deployment cases due to their high locomotive ability in harsh environments such as steep natural terrain with unevenness and obstacles or complex artificial terrain with steps and other movable hazards. The installation of a gripping mechanism at the feet enables the legged robots to climb on steeply inclined slopes, overhanging walls, and even ceilings. Examples of such robots are the Ninja series which can move on glass walls and ceilings~\cite{nagakubo1994}; the Lemur robot series which can traverse rocky cliff faces~\cite{parness2017}; and Magneto, which uses magnets to move along metal structures~\cite{bandyopadhyay2018}.

It is generally not simple to perform experiments with climbing robots because of difficulties accessing harsh natural terrain. In addition, testing often requires large-scale equipment to ensure the safety of the robot and operators. Simulation-based study plays an essential role in developing such a robot that must work reliably in a harsh environment. In the design phase, simulation analysis gives us the fundamental mechanical requirements to decide link geometries, mass distribution, and actuator selection. Also, in the validation phase, simulation in severe or high-risk situations is vital to assess the climbing performance without any physical risks before real-world deployment.

\subsection{Related work}
There have been many physics simulators that have been developed for robot modeling in customizable environments, for example, Gazebo~\cite{koenig2004}, V-Rep~\cite{rohmer2013}, and RaiSim~\cite{hwangbo2018}. The operation and execution of physics calculations of these simulators require relatively little memory due to them being written and supported using mainly C/C++ and Python programs. However, for beginners and middle-level students and researchers, MATLAB has the advantage of the immediate expression of ideas thanks to its intuitive coding style~\cite{fangohr2004}.

The Robotics Toolbox for MATLAB~\cite{corke2017} is one of the most valuable and popular platforms to design and simulate any type of robot, in addition to MathWorks's own Robotic Systems Toolbox. However, these versatile toolboxes support all types of robots and a large variety of libraries. As a result, they are huge, and it can be challenging to understand and familiarize with the whole system.

\subsection{Contribution}
To the authors' best knowledge, simulation platforms focusing purely on legged climbing robots have not yet been established. Therefore, targeting those who want to begin research and development of legged robots with ease and simplicity, we open-sourced the standalone simulation platform: ClimbLab\footnote{Made available at \url{https://github.com/Space-Robotics-Laboratory/ClimbLab}}. ClimbLab is dedicated to legged robots walking and climbing. So far, this simulator depends on only one single toolbox: SpaceDyn~\cite{yoshida1999} for rigid-body dynamics computation, which was also developed and is maintained by the same authors. SpaceDyn is installed automatically with ClimbLab.

This proceedings paper is the initial report to detail the simulation flow of ClimbLab, and also presents some application examples. First, the simulator structure and the computation flow are summarized. Then, representative samples are displayed to show the inspiration of usage.

\section{ClimbLab}
ClimbLab is a standalone simulator integrating all essential aspects necessary for climbing robotics, such as designing robots, simulating various gait strategies under different environmental constraints, and analyzing static and dynamic equilibrium, etc. Almost all functions are usable independent from any MATLAB toolboxes. Instead, dedicated scripts were developed and integrated into a single framework to compute all necessary variables for further analysis of those aspects. The simulator structure and the computational flow are summarized in this section.

\begin{figure}[t]\centering\includegraphics[width=0.86\textwidth]{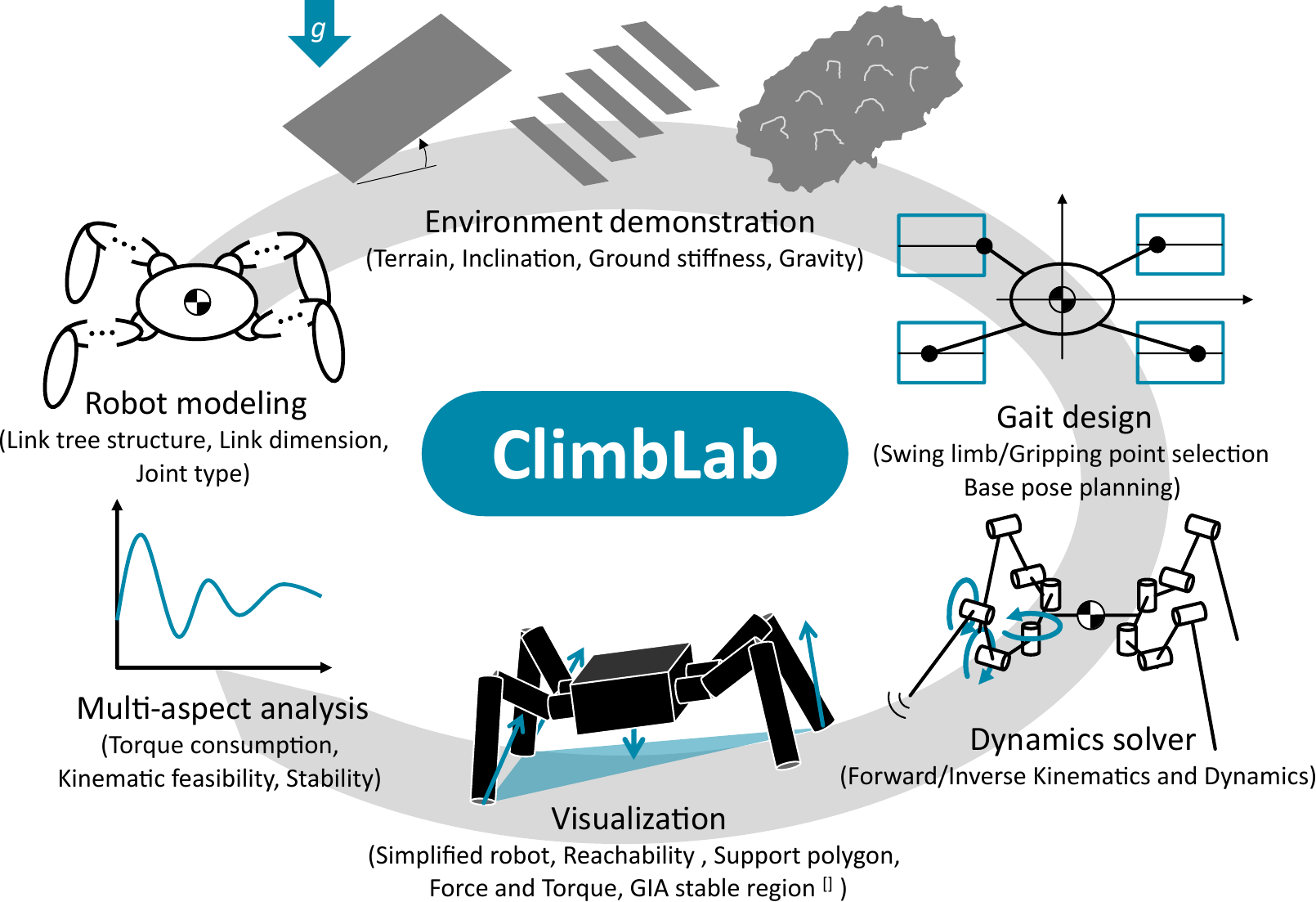}\caption{ClimbLab is a versatile simulation platform for articulated legged robots. The user can first design the robot, set the environmental conditions, and plan the gait strategy. Dynamics calculations are performed by SpaceDyn [9], and measure a variety of variables while the simulation is running. All variables are saved as a single table so that the user can smoothly perform the detailed analysis.}\label{fig:overview}\end{figure}

\begin{figure}[t]\centering\includegraphics[width=0.78\textwidth]{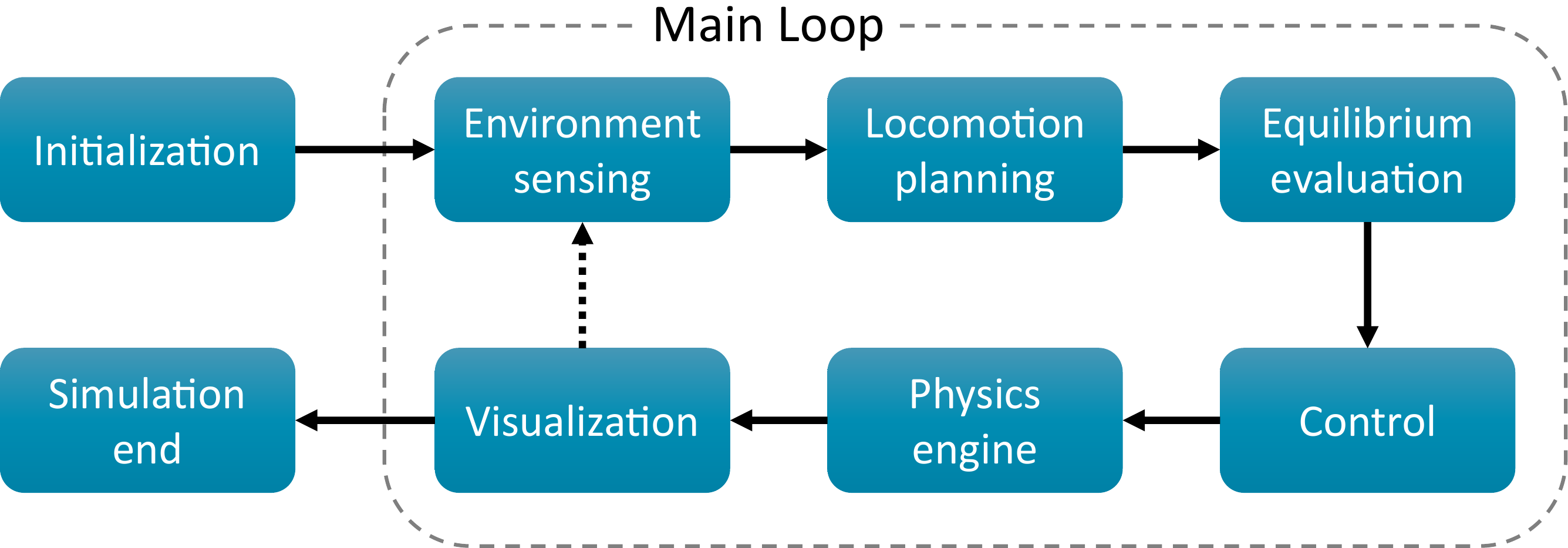}\caption{ClimbLab's each computational phase is composed of a group of subroutine functions, which modularization of the code facilitates the extra development in this platform.}\label{fig:flow}\end{figure}

ClimbLab has one main script file and several function files, divided according to their role inside the simulation. Fig.~\ref{fig:flow} shows a simple diagram for the simulation flow, and further detailed information is described below.

\begin{enumerate}
\item \textit{Initialization}: at the beginning of the simulation, various parameters are defined according to some preset files set by the simulator developers or by arbitrary parameters set by the user.
\begin{enumerate}
\item The robot model is defined according to the SpaceDyn configuration. Currently, insect-type and mammalian-type quadruped robots have been implemented.
\item The ground surface is loaded from a file that contains Cartesian position information. Currently implemented map options are a flat plane, rough terrain generated from fractal structures, and rough terrain extracted from sensed data of a test field made of the sport climbing holds. Other environment parameters can be defined, such as gravitational acceleration, surface inclination, step-time, and simulation time.
\item For periodic gaits, gait parameters can also be initialized, such as step length and height.
\item Files are created to save variables for future analysis.
\end{enumerate}
\item \textit{Environment Sensing}: at the beginning of the main simulation loop, information from the surface is obtained, as it would be on an actual robot collecting point cloud data from a camera. ClimbLab also implements functions to consider a camera field of view to gather only observable data and algorithms to detect graspable point candidates according to the geometry of the surface and the gripping device.
\item \textit{Locomotion planning}: at this section of the simulator, gait planning and local step motion planning are computed.
\begin{enumerate}
\item Different gait types are already implemented in ClimbLab, such as the periodic crawl gait and non-periodic diagonal gaits. The main structure for these gaits is based on the following sequence: checking available grasping points inside the reachable area of each limb, select the next moving limb, select the next point to be grasped, select the next position for the base center of mass (COM), and check kinematic feasibility of next stance. Other gait strategies could also be easily implemented by using this structure.
\item In motion planning, we define how to move each limb or the robot's base by planning the motion trajectory in Cartesian space. Currently, spline trajectories are implemented as an option to connect the initial and final positions. From the desired trajectory and current step time, the current step positions are calculated. Finally, the necessary joint angular positions are computed using analytical inverse kinematics.
\end{enumerate}
\item \textit{Equilibrium Evaluation}: dynamic and static equilibrium of the robot is computed considering the robot's pose, gravity and acceleration effects, and gripping force. Visual plotting of center of gravity and support polygon made by contact points can be used for the classical equilibrium evaluation approach~\cite{mcghee1968}. Additionally, it is possible to compute the tumble stability margin (TSM)~\cite{yoneda1996} and the gravito-inertial acceleration (GIA) margin~\cite{ribeiro2020}. These two methods are recommended for cases of robots with grippers since they also include the effect of the maximum holding force into the tumbling moment.
\item \textit{Control}: the input signal of the actuator of the simulated robot is calculated to achieve the desired joint position. Currently, a proportional and derivative torque control law is performed.
\item \textit{Physics Engine}: the judgment of contact and its resultant ground reaction force is calculated.
\begin{enumerate}
\item The current model for the ground reaction forces is a compliant contact model for all three Cartesian directions, as described in Eq.~(\ref{eq:contact}). In the equation, $\mathbf{F}_r^j$ is the ground reaction force acting at the $j$-th limb end-effector, $\mathbf{x}^j$ is the position of the $j$-th limb end-effector and $\mathbf{x}_c^j$ is its contact point on the surface. The stiffness and damping coefficient matrices for the contact between limb end-effector and surface are $\mathbf{K}_s$ and $\mathbf{D}_s$ respectively. In the case of a robot with grippers, the unilateral condition for contact is not required, so the ground reaction force keeps acting even when the end position is above the surface, as long as the grasping conditions are satisfied. In ClimbLab, we can set the condition for the gripper to detach based on the equilibrium evaluation criteria or on the computed force value and the maximum holding force of the gripper.
\begin{equation}\mathbf{F}_r^j=-\mathbf{K}_s(\mathbf{x}^j-\mathbf{x}_c^j)-\mathbf{D}_s\dot{\mathbf{x}}^j.\label{eq:contact}\end{equation}
\item The equation of motion of a multiple rigid body system with a floating base is solved using the SpaceDyn Toolbox~\cite{yoshida1999}, which is also used to compute forward kinematics. SpaceDyn Toolbox numerically solves the Equation of Motion (EoM) described in Eq.~(\ref{eq:eom}) using the Runge-Kutta method for integration. In the equation, $\mathbf{H}$ is the inertia matrix of the robot, $\mathbf{q}$ is the vector of independent coordinates, including base position and orientation and joints position, $\mathbf{c}_{bm}$ is the term of Coriolis and centrifugal effects. $\mathbf{g}_{bm}$ is the term for gravity effects, $\mathbf{u}$ is the input forces or torques for the joints, $\mathbf{J}$ includes the Jacobian matrices for the base and end-effectors, and $\mathbf{F}_r$ is the term of ground reaction forces.
\begin{equation}\mathbf{H}\ddot{\mathbf{q}}+\mathbf{c}_{bm}+\mathbf{g}_{bm}=\mathbf{u}+\mathbf{J}^{T}\mathbf{F}_r.\label{eq:eom}\end{equation}
\end{enumerate}
\item \textit{Visualization}: simple real-time plots of the robot and surface using MATLAB built-in functions do not require extra toolboxes and do not need a significant computational time for realistic rendering. Additional information can be displayed in the animation, such as support polygons, graspable points, reachable areas, force/acceleration vectors, sensed region, etc. After this section, the simulation iteratively loops back to Sensing until the final step time.
\item \textit{Simulation End}: After the end of the main simulation loop, a file with selectable variables is saved, and time history graphs can be plotted for data analysis.
\end{enumerate}

\section{Application Examples}
We show three representative simulation cases as application examples. First, we show the results of the robot statically standing on an inclined slope. By changing the inclination of the slope, we can find the maximum inclination of the slope that the robot is able to climb.

\subsection{Simulation (1) -- A Robot Holding onto a Cliff}
\begin{figure}[t]\centering\includegraphics[width=0.84\textwidth]{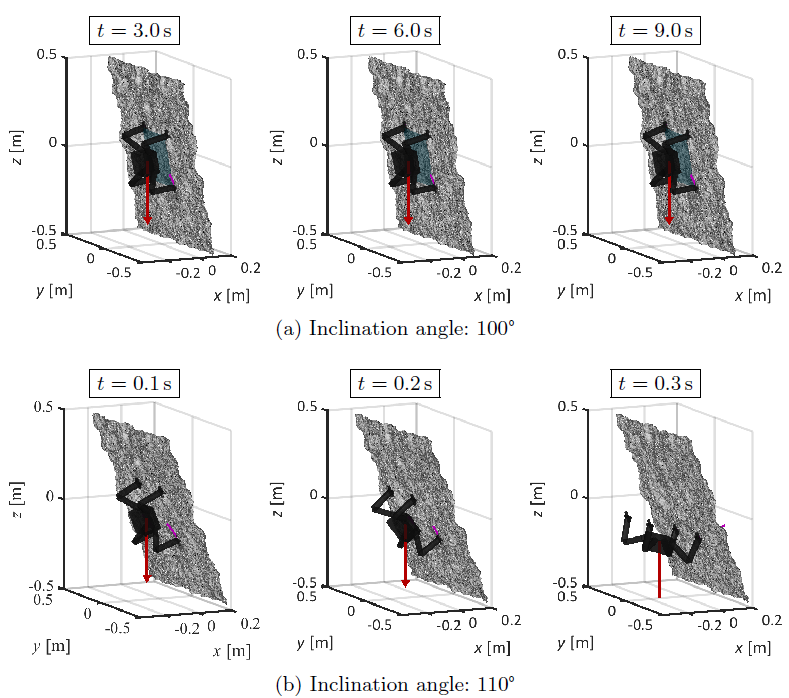}\caption{Snapshots of the simulation results where HubRobo (mass: 2:4 kg) holds onto the overhanging wall in a nominal stance, where inclination is (a) 100° and (b) 110°. This simulation proves the performance of the assumed gripper enables the robot to stand on the 100° inclined wall without losing stability.}\label{fig:hubrobo}\end{figure}

In the first simulation study, a quadrupedal climbing robot testbed HubRobo is demonstrated, supporting the body with its grippers on inclined uneven terrain under Earth's gravitational acceleration. The robot's nominal stance is set as 0.015 m left-right and front-back distance from the base to each foot and 0.08 m of base COM distance from the terrain surface. A gripper that performs 13.0 N of holding force is assumed. We ran the simulation with inclination steps of 10 degrees. Fig.~\ref{fig:hubrobo} shows two cases where the robot attempts to hold on to the overhanging uneven slope inclined at $100^\circ$ and $110^\circ$.

As shown in Fig.~\ref{fig:hubrobo}, the assumed climbing robot system can hold the body on the cliff inclined up to $100^\circ$. However, it fails at $110^\circ$. In other words, we can utilize this simple simulation result to estimate the robot's static performance. Also, by changing the gripping force instead of the environmental settings (terrain inclination and gravity), this type of static simulation can be utilized to investigate the required minimum gripping force for the specific environment expected in a real-world scenario.

\subsection{Simulation (2) -- Gait Analysis for Quadrupedal Robotic Climbing}
Next, we simulate a dog-sized robot climbing on a steep slope. As a representative example of a quadrupedal robot, ANYmal B~\cite{hutter2017} equipped with the grippers at each foot is modeled in ClimbLab. The gripper is assumed to perform 200 N of the adhesion force, and the slope is inclined at $45^\circ$.

In this simulation, we investigate two types of gait. To maximize the velocity while sustaining static stability, a normal gait keeping the stance relative to the base with duty ratio $\beta=0.75$ is selected so that the robot always supports its body with three feet. However, when walking on the steeply inclined ground, this normal gait is not sufficient because consistently translating the robot's base in the forward direction cannot ensure the projection of the COM is kept inside the support polygon. To keep the stability against the tumbling moment and adjust the non-periodic order of the swinging of the legs, which is often required for walking in uneven terrain,~\cite{uno2019} proposed a technique always to place the projection of COM on the intersection of the diagonal lines of the legs. We call this gait a \textit{diagonal gait}. In the diagonal gait, the projection of the COM stays on the edge of the support triangle. Although seemingly at the edge of static stability, the diagonal gait works well since the gripping force at each foot significantly increases the volume of the support polygon, such that the base COM stays well within the stability boundary. In both gaits, the gait period $T=8$ s, so the swing period $(1-\beta)T=2$ s.

\begin{figure}[t]\centering\includegraphics[width=0.82\textwidth]{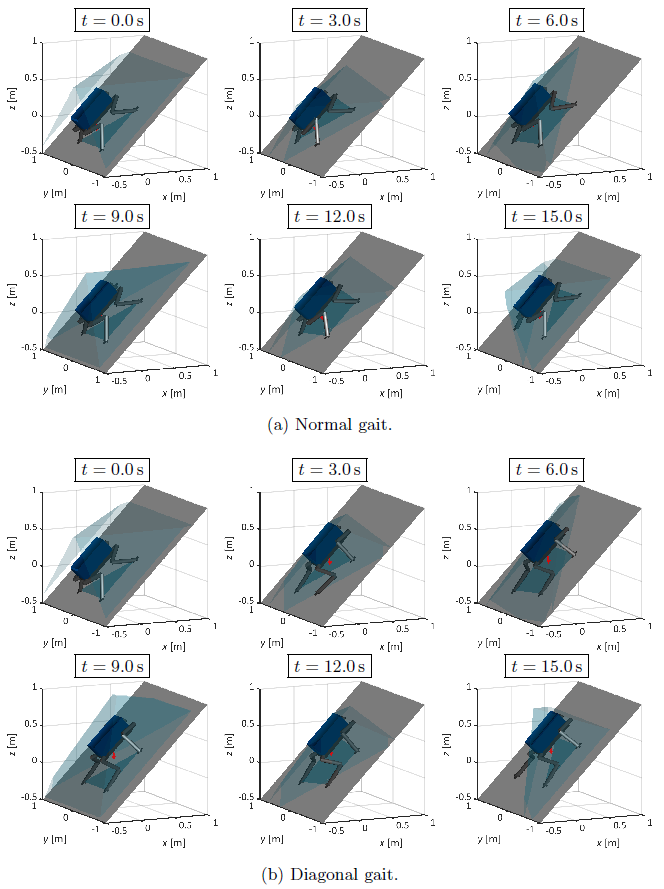}\caption{Snapshots of the simulation results where the robot is climbing on the 45 deg inclined slope in (a) Normal gait and (b) Diagonal gait. GIA vector is drawn as a red arrow and the support polygon and the stable region for GIA [12] are illustrated by the blue-green polygon and polyhedron, respectively.}\label{fig:gaits}\end{figure}

\begin{figure}[t]\centering\includegraphics[width=0.85\textwidth]{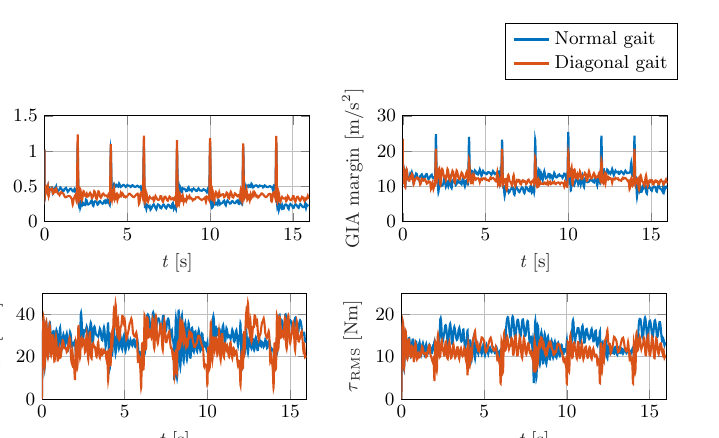}\caption{Time history of multiple metrics to analyze the gait. Two types of stability criterion: TSM [11] and GIA margin [12] in addition to the maximum and RMS of all joints while the simulation, are shown.}\label{fig:metrics}\end{figure}

Fig.~\ref{fig:gaits} shows the comparison of the robot behavior during simulation. Comparing the diagonal gait with the normal gait, we can see that with the diagonal gait, the robot base is moved forward to keep the COM projection on the intersection of the diagonal lines of the leg. Multiple evaluation metrics are plotted during the simulation as shown in Fig.~\ref{fig:metrics}. Stability criterion value, Tumble Stability Margin (TSM)~\cite{yoneda1996}, and Gravito-Inertial Acceleration (GIA) Margin~\cite{ribeiro2020} are supported in ClimbLab. The maximum and the root mean square (RMS) torque of all joints are also tracked in this simulation. As shown in Fig.~\ref{fig:metrics}, while TSM and GIA margins vary depending on which leg is swinging, front or hind, in the normal gait, the diagonal gait exhibits more constant stability.

In addition, statistical values for the metrics are calculated as shown in Table~\ref{tab:metrics}. In terms of stability, there is almost no difference in average value of TSM and GIA margin between the two gaits. However, minimum values are much better in diagonal gait. Also, the diagonal gait uses less torque overall compared to the normal gait. For example, based on the RMS torque, diagonal gait uses 13.5\% less, which results in higher energy efficiency. In summary, the simulator can be utilized for the systematic evaluation of the gait performance.

\begin{table}[t]\caption{Statistical metric values during the simulation case (2)}\label{tab:metrics}\centering
\begin{tabular}{lrr}\hline
Metric & Normal gait & Diagonal gait\\\hline
Mean TSM [m] & 0.361 & 0.359\\
Min. TSM [m] & 0.000 & 0.104\\
Mean GIA margin [m/s$^2$] & 11.8 & 11.9\\
Min. GIA margin [m/s$^2$] & 1.10 & 5.39\\
Mean max. torque of the joints [N m] & 28.7 & 26.7\\
Mean RMS torque of the joints [N m] & 13.3 & 11.5\\\hline
\end{tabular}
\end{table}

\subsection{Simulation (3) -- Sensing and Optimal Footholds Selection}
In the final simulation case, a half-human-sized robot is assumed to walk on the ground where the landable area is restricted, and the robot can contact only the location. As a real robot, ALPHRED~\cite{hooks2020} is modeled as well as the mounted camera: Intel RealSense D435, that is attached to the robot base $60^\circ$ tilted down to sense the front ground.

For the legged robot to explore the rugged terrain, the robot cannot contact everywhere on the ground. This limitation is because there can exist too high obstacles for the robot to pass over and deep ditch that the robot should avoid. Also, parts covered with grass, sand, or moisture are not appropriate to be contacted due to the risk of slipping. To simplify this problem for the legged robots, we restricted the landable positions as random discrete points on the flat plane.

We employed the non-periodic swing limb selection~\cite{uno2020}, and the swing limb's next landing point is determined to maximize the stride to approach the goal position within the kinematic reachability of the limb at each limb swinging phase as presented in~\cite{uno2019}. As for the base motion planner, the diagonal gait introduced in Section~3.2 is chosen. Used gait parameters are as follows: gait period $T=4$ s, so the swing period $(1-\beta)T=1$ s; the constant swing height is 0.05 m. It should be noted that this simulation omitted the dynamics calculation to reduce the simulation time. Kinematic simulation is enough to validate whether the implemented algorithm enables the robot to move forward globally by selecting the optimal local foothold at each swing phase.

\begin{figure}[t]\centering\includegraphics[width=0.74\textwidth]{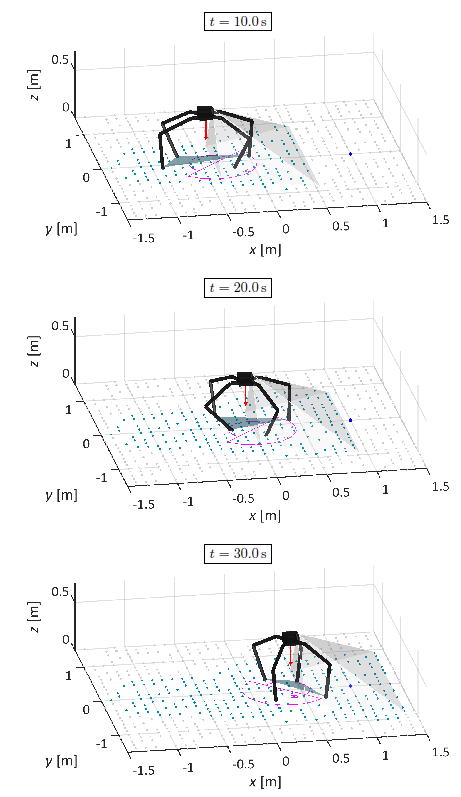}\caption{Snapshots of the final simulation results where the robot is moving forward to the goal point (shown in dark-blue dot at (x; y; z) = (1; 0; 0) [m]) with sensing and selecting the landable points. The sensing camera's field of view is illustrated in light-gray pyramid and the detected footholds are shown in blue-green.}\label{fig:footholds}\end{figure}

The result shown in Fig.~\ref{fig:footholds} indicates the validity of the selection of the optimal local foothold to reach the destination in case the landable points are discrete. In another work, we also presented the efficacy of the non-periodic swing limb selection based on the distribution of the landable points~\cite{uno2020}. In practice for the legged climbing robot, the whole-body motion should be planned, taking into consideration of tumble stability, energy consumption, manipulability in addition to the selectable footholds.

\section{Conclusion}
In this paper, we introduced a MATLAB framework to simulate legged robotic systems walking and climbing in various environments. This simulation platform named ClimbLab is made up of systematically modularized MATLAB functions that facilitate robot design, operation of walking and climbing, and analysis of kinematic and dynamic behavior. Code to reproduce the example simulations shown in this paper is available on the ClimbLab GitHub, where users can quickly build their custom simulation by modifying the reference code. Additional documentation and other simulation examples are also available.

\section*{Acknowledgments}
This work is supported by JSPS KAKENHI Grant Number 19J20685.
\end{document}